\documentclass[11pt,a4paper]{article}
\usepackage[hyperref]{acl2021}
\usepackage{times}
\usepackage{latexsym}

\usepackage{microtype}

\aclfinalcopy 
\def\aclpaperid{2451} 

\usepackage{caption}
\usepackage{subcaption}

\usepackage{graphicx}
\usepackage{booktabs}
\usepackage{multirow}

\usepackage{amssymb}
\usepackage{amsmath}

\usepackage[disable]{todonotes}
\usepackage{url}
\usepackage[capitalise]{cleveref}

\def\tomasz#1{\todo[inline,color=yellow!40]{(tomasz) #1}}

\def\daviddel#1{\bgroup\markoverwith{\textcolor{green}{\rule[0.4ex]{2pt}{3pt}}}\ULon{#1}}

\usepackage[normalem]{ulem}
\def\OD#1{{\color{cyan!80!yellow!80!black!100}OD: \it #1}}
\def\ODdel#1{\bgroup\markoverwith{\textcolor{cyan!89!yellow!80!black!100}{\rule[0.4ex]{2pt}{3pt}}}\ULon{#1}}

\def\remove#1{{\color{gray!70}remove?: \it #1}}
\def\OD#1{\relax}
\def\ODdel#1{#1}

\def\remove#1{\relax}

\newcommand{\orthoprobe}[0]{\emph{Orthogonal Structural Probe}}

\newcommand{\scalingvec}[0]{\emph{Scaling Vector}}

\newcommand{\orthotransf}[0]{\emph{Orthogonal Transformation}}

\title{Introducing Orthogonal Constraint in Structural Probes}
\author{Tomasz Limisiewicz \and David Mare\v{cek} \\
    Institute of Formal and Applied Linguistics, Faculty of Mathematics and Physics \\
    Charles University, Prague, Czech Republic \\
  \texttt{\{limisiewicz, marecek\}@ufal.mff.cuni.cz}
}

\date{}

\begin{document}
\maketitle
\begin{abstract}
 With the recent success of pre-trained models in NLP, a significant focus was put on interpreting their representations. One of the most prominent approaches is structural probing \cite{hewitt-manning-2019-structural}, where a linear projection of word embeddings is performed in order to approximate the topology of dependency structures. In this work, we introduce a new type of structural probing, where the linear projection is decomposed into 1. isomorphic space rotation; 2. linear scaling that identifies and scales the most relevant dimensions. In addition to syntactic dependency, we evaluate our method on novel tasks (lexical hypernymy and position in a sentence). We jointly train the probes for multiple tasks and experimentally show that lexical and syntactic information is separated in the representations.
  Moreover, the orthogonal constraint makes the \emph{Structural Probes} less vulnerable to memorization.
\end{abstract}

\section{Introduction}
\begin{figure}[t]
    \centering
    \begin{subfigure}{0.8\linewidth}
        \centering
        \includegraphics[width=\linewidth]{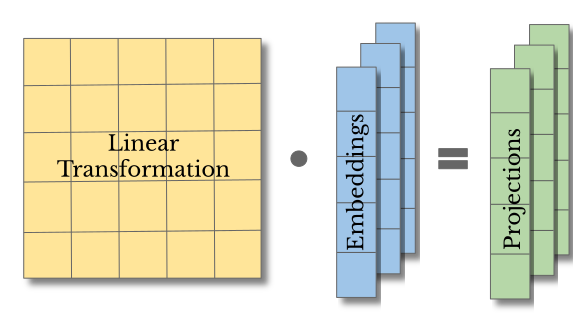}
        \caption{\emph{Structural Probe}}
    \end{subfigure}
    \begin{subfigure}{0.9\linewidth}
        \centering
        \includegraphics[width=\linewidth]{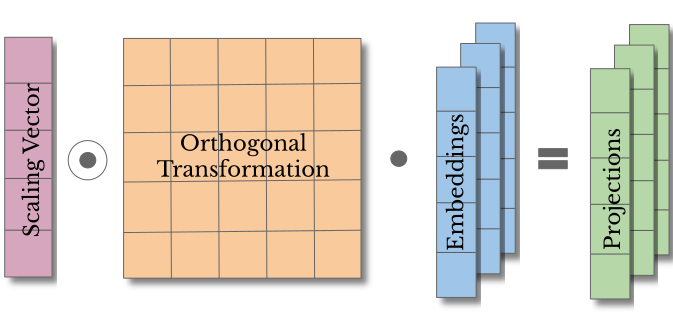}
        \caption{\orthoprobe}
    \end{subfigure}
    
    \caption{Comparison of the \emph{Structural Probe} of \citet{hewitt-manning-2019-structural} and the \orthoprobe~proposed by us.} 
    \label{fig:probes-diagram}
\end{figure}

Latent representations of neural networks encode specific linguistic features. Recently, a lot of focus was devoted to interpret these representations and analyze structures captured by the deep models. One of the most popular analysis methods is probing \cite{belinkov2017neural, blevins2018deep, linzen2016assessing, liu2019linguistic}. The pre-trained model's \footnote{Typically models for language modeling or machine translation are analyzed.} parameters are fixed, and its latent states or outputs are then fed into a simple neural network optimized to solve an auxiliary task, e.g., semantic, syntactic parsing, anaphora resolution, morphosyntactic tagging, etc. The amount of language information stored in the representations can be evaluated by measuring the specific language task's performance.


Probing experiments usually involve classification tasks. Lately, \citet{hewitt-manning-2019-structural} proposed \emph{Structural Probes}, which use regression as an optimization objective. They train a linear projection layer to approximate: 1. dependency tree distances between words\footnote{Tree distance is the length of the tree path between two tokens} by the Euclidean distance between transformed vectors; 2. the tree depth of a word by the norm of its vector.




In Figure~\ref{fig:probes-diagram}, we visualize our \orthoprobe. A linear transformation is replaced by an \orthotransf~(rotation of the embedding space), and product-wise multiplication of rotated vectors by a \scalingvec~to get the final projections. Our motivation is to obtain an embedding space that is isomorphic with the original one, and the impact of each dimension can be evaluated by analyzing \scalingvec's weights. We elaborate on mathematical properties and training details in \cref{sec:ortho-probes}.

In addition to dependency trees used by \citet{hewitt-manning-2019-structural}, we introduce new structural tasks related to lexical hypernymy and word's position in the sentence. We also employ a control task, in which we evaluate the memorization of randomly generated trees.
\orthoprobe \emph{s} let us optimize for multiple objectives jointly by keeping a shared \orthotransf~matrix and changing task-specific \scalingvec\emph{s}.

We will answer the following questions:

\begin{enumerate}
        \item Do our \orthoprobe \emph{s}~achieve comparable or better performance to the \emph{Structural Probes} of \citet{hewitt-manning-2019-structural}?
        \item Finding phenomena such as lexical hypernymy and a word's absolute position in a sentence using \orthoprobe? How vulnerable are the probes to memorizing random data?
        \item Is it possible to effectively train \orthoprobe \emph{s}~jointly for multiple auxiliary objectives, i.e., depth and distance, or multiple types of structures mentioned in the previous question?
        \item Can we identify particular dimensions of the embedding space that encode particular linguistic structures? Are there any superfluous dimensions?
        \item If yes, what is the relationship between subspaces encoding distinct structures?
\end{enumerate}

\section{Related Work}
Basic linguistic features can be easily extracted from the contextual representations \cite{liu2019linguistic}. Probing was intensively used to investigate the representation of morphological information (mainly POS tags) in hidden states of machine translation systems and language models \cite{belinkov2017neural, peters2018deep, tenney-2018-learn}. Besides the work of \citet{hewitt-manning-2019-structural}, probing for dependency syntax was performed by \citet{tenney-2019-bert} and \citet{blevins2018deep}. They utilize a binary classifier to predict dependency edges. 
In work contemporary to ours, \citet{ravichander-2020-systematicity} employ a softmax classifier to show that  BERT can be successfully probed for hypernymy. 

There is an ongoing debate on which probe architectures offer a good insight into underlying representations. \citet{zhang-bowman-2018-language} showed that a POS tagger on top of a frozen randomly initialized LSTM model achieves unexpectedly high results. In the work of \citet{hewitt-liang-2019-designing}, the multilayer perceptron probes display similar accuracy for predicting POS tags as for randomly assigned tags. These symptoms underscore how crucial it is to carefully consider the probe's architecture to avoid reaching spurious conclusions. It is good practice to monitor additional aspects of the probe beyond performance on a linguistic task, such as selectivity \cite{hewitt-liang-2019-designing}, or complexity \cite{pimentel-2020-pareto}. The recent state of knowledge is summarized in surveys on probing \cite{belinkov2019analysis} and interpretation of BERT's representations \cite{rogers-2020-primer}.

\paragraph{Orthogonality}
\phantomsection
has been applied broadly in the field of deep learning, especially to cope with exploding/vanishing gradient problem in recurrent neural networks \cite{arjovsky-2016-unitary, jing-2017-gated, wisdom-2016-fullcapacity}.
In this work, we use regularization to enforce the orthogonality of a dense layer. In literature, such an approach is called ``soft constraint'' \cite{bansal-2018-can, vorontsov-2017-orthogonality}. Alternatively, ``hard constraint'' assumes parameterization of a network such that the transformation of latent states is orthogonal by definition \cite{arjovsky-2016-unitary, jing-2017-eunn}. There are a few examples of orthogonality applications in NLP: in RNN language model \cite{dangovski-2019-rotational}; in Performer \cite{choromanski-2020-rethinking}, which is a more efficient counterpart of Transformer \cite{vaswani2017attention}.
Best to our knowledge, we are the first to use orthogonal transformation in probing.

\section{Method}
\label{sec:ortho-probes}


In this section, we first review the structural probing proposed by \citet{hewitt-manning-2019-structural} and then introduce our \orthoprobe. 

\subsection{Structural Probes}

In the previous work, a linear transformation is optimized to transform the contextual word representations
produced by a pre-trained neural model (e.g. BERT \citet{devlin2019bert}, ELMo \citet{peters2018deep}). 
The squared L2 norm of the differences between transformed word vectors approximate the tree distance between them:
\begin{equation}
    \label{eqn:distance-probe}
    d_B(h_i,h_j)^2 = (B(h_i - h_j))^T(B(h_i - h_j)),
\end{equation}
where $B$ is the \emph{Linear Transformation} matrix and $h_i$, $h_j$ are the vector representations of words at positions $i$ and $j$.

The probe is optimized to approximate the distance between tokens in the dependency tree ($d_T$) by gradient descent objective:
\begin{equation}
    \min_{B}\frac{1}{s^2}\sum_{i,j}\bigl\lvert d_T(w_i,w_j) - d_B(h_i, h_j)^2\bigr\rvert,
\end{equation}
where $s$ is the length of a sentence.

Moreover, the same work introduced depth probes, where vectors were linearly transformed so that the squared L2 length of the mapping approximate the token's depth in a dependency tree:
\begin{equation}
    \label{eqn:depth-probe}
    ||h_i||_B^2 = (Bh_i)^T(Bh_i)
\end{equation}
Gradient descent objective is analogical:
\begin{equation}
    \min_{B}\frac{1}{s}\sum_{i}\bigl\lvert  \lVert w_i \rVert_T - \lVert h_i \rVert_{B}^2\bigr\rvert
\end{equation}

\subsection{Orthogonal Structural Probes}

We introduce orthogonality to structural probes. For that purpose, we perform the singular value decomposition of the matrix $B$
\begin{equation}
    B = U \cdot D \cdot V^T,
\end{equation}
where the matrices $U$ and $V$ are orthogonal
, and $D$ is diagonal. Notably, when we substitute $B$ with $U\cdot D\cdot V^T$ in \cref{eqn:distance-probe}, the matrix $U$ cancels out. It can be easily shown by rearranging the variables in the equation:\footnote{A complete derivation can be found in the appendix.}
\begin{equation}
\label{eqn:orthogonal-probe-derivation}
\begin{split}
    &d_B(h_i,h_j)^2 \\
    &= (D V^T (h_i - h_j))^T(D V^T (h_i - h_j))
\end{split}
\end{equation}

We can replace the diagonal matrix $D$ with a vector $\bar{d}$ and use element-wise product (we will call $\bar{d}$ the \scalingvec). Finally, we get the following equation for \emph{Orthogonal Distance Probe}:
\begin{equation}
\begin{split}
    &d_{\bar{d}V^T}(h_i,h_j)^2 \\
    &=(\bar{d} \odot V^T  (h_i - h_j))^T( \bar{d} \odot V^T (h_i - h_j))
\end{split}
\end{equation}
The same reasoning can be applied to \cref{eqn:depth-probe} to obtain \emph{Orthogonal Depth Probe}:
\begin{equation}
    ||h_i||_{\bar{d}V^T}^2 = (\bar{d} \odot V^T  h_i)^T( \bar{d} \odot V^T h_i)
\end{equation}

We showed that \orthoprobe~is mathematically equivalent to \emph{Standard Structural Probe}.


\subsection{Multitask Training}
\label{sec:multitask}

\orthoprobe~can be easily adapted to multitask probing for a set of objectives $\mathcal{O}$. We use one shared \orthotransf~and different \scalingvec\emph{s} for each task. In one batch, we compute a loss for a specific objective. For each batch (with objective $o \in \mathcal{O}$), a forward pass consists of multiplication by a shared orthogonal matrix $V^T$ and product-wise multiplication by a designated vector $\bar{d}_o$. All the batches are shuffled together in a training epoch. 

\subsection{Orthogonality Regularization}

We use \emph{Double Soft Orthogonality Regularization} (DSO) proposed by \citet{bansal-2018-can} to coerce orthogonality of the matrix $V$ during training:
\begin{equation}
    \label{eqn:dso}
    \lambda_O DSO(V) = \lambda_O(||V^TV - \mathbb{I}||^2_F + ||VV^T - \mathbb{I}||^2_F)
\end{equation}
$||\cdot||_F$ stands for the Frobenius norm of a matrix.

\subsection{Sparsity Regularization}

In further experiments, we investigate the effects of sparsity in \scalingvec. For that purpose, we compute the L1 norm and add it to the training loss.
\begin{equation}
    \label{eqn:L1}
    \lambda_{S} \lVert \bar{d} \rVert_1
\end{equation}

\subsection{Training Objective}

Altogether, the loss equation in \emph{Orthogonal Distance Probe} for objective $o \in \mathcal{O}$ is the following:
\begin{equation}
\begin{split}
    \label{eqn:orthogonal-distance-probe-loss}
    L_{o,dist.} = \frac{1}{s^2}\sum_{i,j}\bigl\lvert d_T(w_i,w_j) - d_{\bar{d_o}V^T}(h_i, h_j)^2\bigr\rvert+ \\ 
    + \lambda_O DSO(V) + \lambda_{S} \lVert \bar{d_o} \rVert_1
\end{split}
\end{equation}
And in \emph{Orthogonal Depth Probe}:
\begin{equation}
\begin{split}
    \label{eqn:orthogonal-depth-probe-loss}
    L_{o,depth} = \frac{1}{s}\sum_{i}\bigl\lvert \lVert w_i \rVert_T - \lVert h_i \rVert_{\bar{d_o}V^T}^2 \bigr\rvert+ \\ 
    + \lambda_O DSO(V) + \lambda_{S} \lVert \bar{d_o} \rVert_1
\end{split}
\end{equation}
The loss is normalized by the number of predictions in a sentence and averaged across a batch.





\section{Experiments}
We train probes on top of each of 24 layers of English BERT large cased model \cite{devlin2019bert} implemented by HuggingFace \citep{wolf-2020-transformers}. We optimize for the approximation of depth and distance in four types of structures: syntactic dependency, lexical hypernymy, absolute position in a sentence, and randomly generated trees.
In the following subsection, we expand upon these structures.

\subsection{Data and Objectives}

In our experiments, we use training, evaluation, and test sentences from Universal Dependencies English Web Treebank \cite{silveira-2014-gold}. Depending on the objective, we reveal only partial relevant annotation from the dataset.

\OD{(make sure you say which is your new addition and what is taken over from H+M, maybe add a picture with examples for all?)}
\paragraph{Dependency Syntax}
\phantomsection
We probe for syntactic structure in Universal Dependencies parse trees \cite{nivre-etal-2020-ud}. Dependency trees are annotated in English Web Treebank. We focus on distances between words in dependency trees and their depth, i.e., distance from the syntactic root.
\paragraph{Lexical Hypernymy}
\phantomsection
We introduce probing for lexical information. We optimize probes to approximate the distance between pairs of words in the hypernymy tree and the depth for each word. For that purpose, we use the tree from WordNet \cite{miller-1995-wordnet}. We consider lexical distances 
between pairs of nouns and pairs of verbs in sentences and lexical depth for each noun and verb. 
We provide gold POS information and look up synset by a lemmatized form of a word to avoid ambiguity.


\paragraph{Position in a Sentence}
\phantomsection

Probing for the sentence index of a word and positional difference between pairs of words.

\paragraph{Random Structures}
\phantomsection

We probe for randomly generated trees. When we jointly optimize for depth and distance, we keep the same randomly generated tree. This control task allows us to determine the extent to which our probes memorize the structures and thus over-fit to the training data.


\subsection{Training}

We use batches of size $12$ and an initial training rate of $0.02$. We use learning rate decay and early-stopping mechanism: if validation loss does not achieve a new minimum after an epoch, the learning rate is divided by $10$. After three consecutive learning rate updates not resulting in a new minimum, the training is stopped.
\paragraph{Orthogonality Regularization}
\phantomsection
In our experiments, we took $\lambda_{O}$ equal to $0.05$.\footnote{We experimentally checked that ten times smaller and ten times larger values of $\lambda_{O}$ do not affect orthogonality of matrix $V$ and lead to the same results.} The regularization converged early during the gradient optimization. Hence we can assume that matrix $V$ is orthogonal. 
\paragraph{Sparsity Regularization}
\phantomsection
By default $\lambda_{S}=0$. Only in the experiments described in \cref{sec:dim-selection-results}, we use sparsity regularization by setting $\lambda_{S}$ to a positive value ($0.005$, $0.05$, or $0.1$) when DSO drops below $1.5$ during the training. This mechanism prevents weakening orthogonality constraint in early epochs.

Additional details of the training are described in the appendix. The code is available at GitHub: \url{https://github.com/Tom556/OrthogonalTransformerProbing}.

\subsection{Evaluation}

We assess Spearman's rank correlation between gold and predicted values. We report the average correlations for the sentences with lengths from 5 to 50 in the same way as \citet{hewitt-manning-2019-structural}.

Our \orthoprobe \emph{s} are trained jointly for multiple objectives (\cref{sec:multitask}). We evaluate the effect of multitasking testing different configurations: \textbf{A)} separate probing for each objective; \textbf{B)} joint probing for distance and depth in the same structure type; \textbf{C)} joint probing for distance in all structures; \textbf{D)} joint probing for depths in all structures; \textbf{E)} probing for all objectives together.
We compare the results with two baselines: \textbf{I)} optimizing only \scalingvec; \textbf{II)} \emph{Structural Probes}.

\subsection{Dimensionality of Scaling Vector}
\label{sec:dim-selection-description}


We hypothesize that the orthogonality regularization allows us to find embedding subspace capable of representing a particular linguistic structure. \remove{In \orthotransf, vectors are projected onto that subspace.} In \cref{sec:dim-selection-results}
, we examine the performance of lower-rank projections and ask whether further restrictions of dimensionality affect the results. In \cref{sec:separtation} we analyze interactions between subspaces related to a particular objective in a joint probing setting.

\begin{table*}[!ht]
\centering
\small
\begin{tabular}{@{}l|c|c|c|ccc@{}}
\toprule
 & \textbf{I} & \textbf{II} & \textbf{A} & \textbf{B} & \textbf{C} / \textbf{D} & \textbf{E} \\ \cline{2-7}
 & & & & \multicolumn{3}{c}{multitask orthogonal probing} \\ \cline{5-7} 
  &
  \begin{tabular}[c]{@{}c@{}} Scaling\\ Vector \\only \end{tabular} &
  \begin{tabular}[c]{@{}c@{}}Structural\\ Probe\end{tabular} &
  \begin{tabular}[c]{@{}c@{}}Orthogonal\\ Structural\\ Probe\end{tabular} &
  \begin{tabular}[c]{@{}c@{}}distance \\+ depth \end{tabular} &
  \begin{tabular}[c]{@{}c@{}}all distances\\ or all depths\end{tabular} &
  all tasks \\ \hline
DEP Depth   & .459 $_{\pm.001}$ & .856 $_{\pm.001}$ &  \underline{\bf.858} $_{\pm.001}$  & .855 $_{\pm.001}$  & .850 $_{\pm.002}$ & .852 $_{\pm.001}$ \\
Layer  & 17 & 18 & 17    & 16    & 16    & 16    \\ \midrule
DEP Dist. & .513 $_{\pm.001}$ & \underline{\bf.843} $_{\pm.001}$ & \bf.842 $_{\pm.001}$  & .838 $_{\pm.001}$ &  .833 $_{\pm.001}$ & .832 $_{\pm.002}$ \\
Layer &  18 & 17    & 17    & 17    & 17    & 16    \\ \midrule
LEX Depth & .572 $_{\pm.001}$ & \underline{\bf.892} $_{\pm.002}$ & .882 $_{\pm.002}$ & .869 $_{\pm.005}$ & .885 $_{\pm.004}$ & .873 $_{\pm.005}$  \\
Layer  & 13 & 8     & 8     & 8     & 6     & 9     \\ \midrule
LEX Dist.  &  .560 $_{\pm.001}$ & \underline{\bf.816} $_{\pm.008}$ & .803 $_{\pm.005}$ &  .789 $_{\pm.004}$ & .792 $_{\pm.010}$ & .792 $_{\pm.005}$\\
Layer  & 13 & 6     & 6     & 7     & 6     & 6     \\ \midrule
POS Depth   & .232 $_{\pm.013}$ & \underline{\bf.989} $_{\pm.001}$ & .983 $_{\pm.001}$ & .986 $_{\pm.001}$ & .976 $_{\pm.004} $ & .982 $_{\pm0.001}$\\
Layer  & 5 & 1     & 6     & 1     & 2     & 3     \\ \midrule
POS Dist.  & .441 $_{\pm 0.001} $ & \underline{\bf.980} $_{\pm.001}$ & .979 $_{\pm.001}$ & .977 $_{\pm .001}$  & .978 $_{\pm.001}$ & .976 $_{\pm0.001}$ \\
Layer  & 1 & 4     & 4     & 4     & 5     & 4     \\ \midrule
RAND Depth & .008 $_{\pm .002}$ & .206 $_{\pm.010}$ & .136 $_{\pm.007}$ & .129 $_{\pm .010} $ & .163 $_{\pm.023}$ & \underline{\bf.107} $_{\pm.019}$ \\
Layer  &  6 & 17    & 18    & 18    & 18    & 19    \\ \midrule
RAND Dist. & .149 $_{\pm.001}$ & .242 $_{\pm.005}$ & .220 $_{\pm.006}$ & \underline{\bf.206} $_{\pm.004}$  & \bf.209 $_{\pm.005}$ & \bf.208 $_{\pm.007}$ \\
Layer  & 17 & 19    & 18    & 17    & 19    & 15    \\ \midrule
AVG. DEP, LEX, POS  & .463 & \underline{.896} & .891 & .886 & .886 &  .883 \\ 
ABOVE - AVG. RAND   & .385 & .673 & .713 & .718 & .699  & \underline{.726} \\  \bottomrule
\end{tabular}
\caption{The highest Spearman's  correlations (across layers) between predicted values and gold annotations on a held out test set (for random structures computed on a train set). Each column represents another variant of training. Standard deviation was calculated for six runs. \underline{Each row's optimal result} is underlined (except baseline \textbf{I}); \textbf{results within 95\% confidence interval} based on Student's t-test \cite{student-1908-probable} are marked in bold.} 
\label{tab:correlation}
\end{table*}

\begin{figure}[!h]
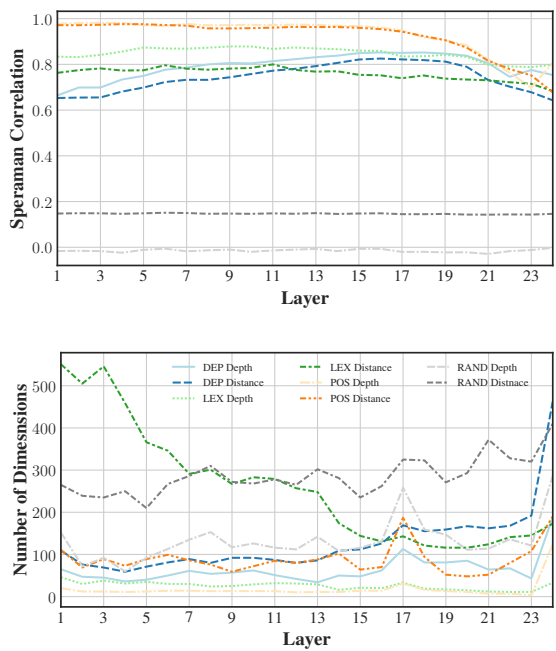

    \centering
    \begin{subfigure}{1.0\linewidth}
        \centering
        \resizebox{1.0\linewidth}{!}{\input{figures/bert.large.spearman.new.pgf}}
    \end{subfigure}
        \begin{subfigure}{1.0\linewidth}
        \centering
        \resizebox{1.0\linewidth}{!}{\input{figures/bert.large.ndimensions.pgf}}
    \end{subfigure}
    \caption{Spearman's correlations and number of non-zero \scalingvec's dimensions across layers for joint training.}
    \label{fig:acrosslayers}
\end{figure}

\section{Results}
\label{sec:results}
\begin{table*}
\centering
\begin{tabular}{@{}l|cc|ccc|cc|cc|cc@{}}
\toprule
\multirow{3}{*}{} &
  \multicolumn{2}{c}{Subspace}  &
  \multicolumn{3}{|c}{Share of Dropped} &
  \multicolumn{6}{|c}{Sparsity Regularization}\\ 
              & & &
              \multicolumn{3}{c}{Dimensions} &
              \multicolumn{2}{|c}{$\lambda_S=0.005$} & \multicolumn{2}{|c}{$\lambda_S=0.05$} & \multicolumn{2}{|c}{$\lambda_S=0.1$} \\
              & Dims & Corr & 25\% & 33\% & 50\% & Dims & Corr & Dims & Corr & Dims & Corr\\ \midrule
DEP Depth     &  
137 & .858 & .783 & .758 & .700 & 26 & .856 & 2 & .832 & 1 & .822\\
DEP Dist.     & 
189 & .842 & .800 & .781 & .741 & 76 & .835 & 21 & .784 & 14 & .746 \\ \midrule
LEX Depth     & 
19  & .884 & .841 & .822 & .784 & 19 & .875 & 11 & .852 & 10 & .836 \\
LEX Dist.     & 
263 & .805 & .768 & .755 & .722 & 92 & .792 & 60 & .756 & 52 & .737 \\ \midrule
POS Depth     & 
20  & .983 & .760 & .686 & .526 & 11 & .982 & 6 & .981 & 3 & .981 \\
POS Dist.     & 
98  & .979 & .890 & .859 & .627 & 38 & .978 & 14 & .975 & 11 & .970 \\ \midrule
RAND Depth    & 
259 & .128 & .108 & .101 & .091 & 6 & .037 & 1 & .011 & 1 & .010 \\
RAND Dist.    & 
399 & .222 & .215 & .213 & .208 & 116 & .208 & 20 & .163 & 13 & .155 \\ \bottomrule
\end{tabular}
\caption{The highest Spearman's  correlations (across layers) between predicted values and gold annotations on a held-out test set (for random structures computed on a train set). In columns 2-3,  results, when only selected dimensions are used. In columns 4-6, a portion of the selected dimensions is masked. In columns 7-12, sparsity regularization with different $\lambda_S$ is applied. Probing for one objective.}
\label{tab:subspace}
\end{table*}

We compare Spearman's correlations between predicted values and gold tree depths and distances in \cref{tab:correlation}. The correlations obtained from \orthoprobe \emph{s} are high for linguistic structures: from $0.803$ for lexical distance to $0.882$ for lexical depth. Predicted positional depths and distances nearly match gold values.

Correlation on training data for random structures is very weak, hinting that the probes do not memorize structures during training but extract them from the model's representations. The correlation for distances is higher than for depth. We hypothesize it is because the probes learn some basic tree properties.\footnote{For instance, when the distances between nodes X and Y, and Y and Z are both 1, then the distance between X and Z needs to be 2}

The results obtained by \orthoprobe \emph{s}~are close to those of \emph{Structural Probes}. For dependency distance, the difference is not statistically significant. Notably, correlations on training set for randomly generated trees decreased. It suggests that \orthoprobe \emph{s} are less vulnerable to memorization. In multitask probing, 
correlation evenly decreases across all tasks. While selectivity (the difference between average correlation for dependency, lexical, and positional objectives and random objectives) increases from $0.673$ to $0.726$.
Optimizing only a \scalingvec~gives distinctly lower correlations. These results emphasize the necessity of changing the coordinate system to amplify the dimensions encoding linguistic information.



In \cref{fig:acrosslayers} (upper), we observe that the performance varies throughout the layers, confirming previous observations by \citet{hewitt-manning-2019-structural} and \citet{tenney-2019-bert}. The mid-upper layers tend to be more syntactic, and the mid-lower ones are more lexical. Predicting word position is more accurate in the lower layers, dropping significantly toward the last layers. It is due to the fact that in BERT, positional embeddings are added before the first layer. Random structure probes maintain steady results across all the layers.




\subsection{Dimensionality}
\label{sec:dim-selection-results}

We observe that orthogonality constraint is quite effective in restricting the probe's rank. 
In most of our experiments, the majority of \scalingvec ~parameters converged to zero. It allows selecting subspaces encoding particular linguistic features. We want to answer whether such subspace has enough capacity for each probing task. For that purpose, we zero out the dimensions with corresponding \scalingvec~weights closer to zero than $\epsilon=10^{-4}$.\footnote{In the appendix, we show that dimension selection is not sensitive to the selection of low $ 10^{-30} < \epsilon < 10^{-3}$.}
Their elimination does not affect the results; correlations in \cref{tab:subspace} and \cref{tab:correlation} column \textbf{A} are practically equal. The dimensionality reduction is the strongest for lexical and positional depth probes, where subspaces with the rank of 19 and 20 respectively encode the structures as well as the whole embedding space with 1024 dimensions (\cref{fig:acrosslayers}, lower). The number of selected dimensions is the highest in probing for random structures. This is because a large capacity is required for memorization.

Another question we pose is whether it would be adequate to shrink the subspace even further. For each objective, we choose and drop a random portion of parameters to examine how it would affect the predictions. We conduct a procedure similar to cross-validation, i.e., we repeatedly drop disjoint and exhaustive sets of dimensions and average results for each set at the end.\footnote{When we drop $25\%$ of dimensions, we randomly choose four sets. Each dimension is exactly in one set.} \cref{tab:subspace} shows that dimension dropping had the largest impact on positional probes: $-0.458$ for depth; the decrease is low for lexical distance -- only $-0.083$. 
It suggests that the information necessary for the latter objective is more dispersed than for the former one.

\begin{table}[!t]
\centering
\scriptsize
    \begin{tabular}{ll|cc|cc|cc|cc}
    & & \multicolumn{2}{c}{DEP} & \multicolumn{2}{|c}{LEX} & \multicolumn{2}{|c}{POS} & \multicolumn{2}{|c}{RAND} \\
    & & \rotatebox{90}{Depth} & \rotatebox{90}{Dist.} & \rotatebox{90}{Depth} & \rotatebox{90}{Dist.} & \rotatebox{90}{Depth} & \rotatebox{90}{Dist.} & \rotatebox{90}{Depth} & \rotatebox{90}{Dist.} \\\hline
    \multirow{5}{*}{\rotatebox{90}{DEP}} &  &  &   &   &  &  &  &  & \\
    & Depth & 62 & 48  & 0  & 0   & 10 & 19 & 23  & 21  \\
     & &  &   &   &  &  &  &  & \\
    & Dist. &  & 126 & 0  & 0   & 9  & 23 & 25  & 30  \\ 
     & &  &   &   &  &  &  &  & \\ 
    \hline
    \multirow{5}{*}{\rotatebox{90}{LEX}} &  &  &   &   &  &  &  &  & \\
    & Depth &  &   & 20 & 18  & 0  & 4  & 1   & 5   \\
    & &  &   &   &  &  &  &  & \\
    & Dist. &  &  & & 131 & 0  & 7  & 5   & 19  \\ 
    & &  &   &   &  &  &  &  & \\ \hline
    \multirow{5}{*}{\rotatebox{90}{POS}} &  &  &   &   &  &  &  &  & \\
    & Depth &  &   &  &   & 14 & 10 & 13  & 10  \\
    & &  &   &   &  &  &  &  & \\
    & Dist. &  &   &  &   & & 70 & 33  & 50  \\
    & &  &   &   &  &  &  &  & \\ \hline
    \multirow{5}{*}{\rotatebox{90}{RAND}} &  &  &   &   &  &  &  &  & \\
    & Depth &  &  &  &   & & & 131 & 95  \\
    & &  &   &   &  &  &  &  & \\
    & Dist. &  &   &  &   &  &  &  & 262 \\
    & &  &   &   &  &  &  &  & \\ \hline
    \end{tabular}
    \caption{The number of shared dimensions selected by \scalingvec~after the joint training of probe on top of the 16th layer.}
    \label{tab:l-16-separation}
\end{table}

\paragraph{Sparsity Regularization}
We use sparsity regularization of \scalingvec~to examine whether dimensionality can be reduced more intelligently. The strength of regularization is regulated by value of $\lambda_s \in \{0.005, 0.05, 0.1\}$. We observe that for some objectives (dependency depth, positional depth, and positional distance), the relevant information is captured in a small number of dimensions. Remarkably, only one dimension of embedding space can achieve $0.822$ correlation with dependency depths. We conjecture that if it is possible to achieve a high correlation with sparse subspaces, information on the phenomenon is focal in the model (concentrated in few dimensions). For the objectives with focal information, results decrease sharply when random dimensions are dropped because the probability of dropping important coordinates is high. On the other end of the spectrum, we can identify the objective for which information is spread -- lexical distance. The dropping of random dimensions only moderately decreases correlation, as there are no especially essential coordinates. Probing with sparsity regularization produces subspaces of relatively large size.

Sparsity regularization also positively affects control objectives, decreasing correlations with distances and depths of randomly generated structures, indicating that regularized probes are less prone to memorization.

Notably, \citet{hennigen-2020-intrinsic} proposed a method for selecting embeddings' dimensions relevant to particular linguistic phenomena. In our setting, thanks to the \orthotransf, we are not constrained to analyzing the dimensions of just one coordinate system.

\subsection{Separation of Information}
\label{sec:separtation}


Another outcome of joint training \remove{and dimensionality reduction} was the ability to examine relationships between subspaces for each of the objectives. Figure~\ref{fig:dep-lex-separation} shows histograms of the dimensions selected in lexical and dependency probes. Each bin of the histogram corresponds to 10 coordinates. The height of a bar (in one color) represents how many were selected for a specific task. The dimensions on the x-axis are ordered by the weighted
absolute values of \scalingvec \emph{s}.\footnote{We weight the values before sorting to keep together non-zero dimensions of each \scalingvec, i.e., dependency depth values are multiplied by 1000, dependency distance 100, lexical depth by 10. The weighting is performed only for visualization; the separation of linguistic information can be observed independently in \cref{tab:l-16-separation}.} 

We found that in layers 6 and 16 (they achieve the highest correlation in lexical and dependency, respectively), the histograms are disjoint, indicating that the layers' representations of dependency syntax and lexical hypernymy are orthogonal to each other in the embedding space. The orthogonality is less visible in the first layer and disappears almost entirely in the top one. In most layers, depth subspace is included in distance subspace for the same structural type. This behavior was expected as distance probing is more complex and therefore requires more capacity.

\begin{figure}[!ht]
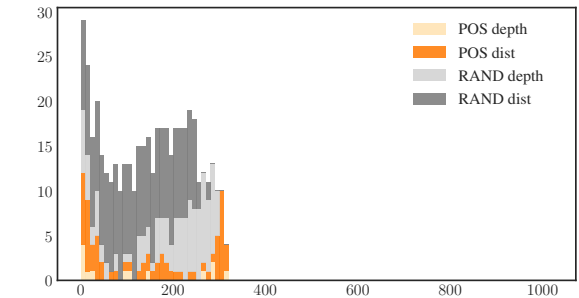

    \centering
        \begin{subfigure}{0.99\linewidth}
        \centering
        \resizebox{1.0\linewidth}{!}{\input{figures/disentanglement/hist_dep_depth_dep_distance_pos_depth_pos_distance_layer_16_legend.pgf}}
    \end{subfigure}
    \begin{subfigure}{0.99\linewidth}
        \centering
        \resizebox{1.0\linewidth}{!}{\input{figures/disentanglement/hist_dep_depth_dep_distance_rnd_depth_rnd_distance_layer_16_legend.pgf}}
    \end{subfigure}
        \begin{subfigure}{0.99\linewidth}
        \centering
        \resizebox{1.0\linewidth}{!}{\input{figures/disentanglement/hist_lex_depth_lex_distance_pos_depth_pos_distance_layer_16_legend.pgf}}
    \end{subfigure}
        \begin{subfigure}{0.99\linewidth}
        \centering
        \resizebox{1.0\linewidth}{!}{\input{figures/disentanglement/hist_lex_depth_lex_distance_rnd_depth_rnd_distance_layer_16_legend.pgf}}
    \end{subfigure}
        \begin{subfigure}{0.99\linewidth}
        \centering
        \resizebox{1.0\linewidth}{!}{\input{figures/disentanglement/hist_pos_depth_pos_distance_rnd_depth_rnd_distance_layer_16_legend.pgf}}
    \end{subfigure}
    \caption{Histograms of dimensions selected by \scalingvec~after the joint training of probe on top of the 16th layer. Best in color.}
    \label{fig:l-16-separation}
\end{figure}

In \cref{fig:l-16-separation} we present histograms for additional tasks at the model's 16th layer. The positional subspace has a sizable intersection with the syntactic one, yet only a few common dimensions with the lexical subspace. The connection can be attributed to the fact that dependency edges can often be inferred from words' relative positions. Probing for random structures is interlinked with other objectives. \OD{(randomly? maybe try to make it clearer that it's to be expected?)} The sizes of shared subspaces for each pair can be found in \cref{tab:l-16-separation}. Histograms and tables for other sets of tasks are presented in the appendix.

\begin{figure}[!ht]
    \centering
    \begin{subfigure}{0.99\linewidth}
        \centering
        \resizebox{1.0\linewidth}{!}{\input{figures/disentanglement/hist_lex_depth_lex_distance_dep_depth_dep_distance_layer_1.pgf}}
    \end{subfigure}
    \begin{subfigure}{0.99\linewidth}
        \centering
        \resizebox{1.0\linewidth}{!}{\input{figures/disentanglement/hist_lex_depth_lex_distance_dep_depth_dep_distance_layer_6.pgf}}
    \end{subfigure}
        \begin{subfigure}{0.99\linewidth}
        \centering
        \resizebox{1.0\linewidth}{!}{\input{figures/disentanglement/hist_lex_depth_lex_distance_dep_depth_dep_distance_layer_16.pgf}}
    \end{subfigure}
        \begin{subfigure}{0.99\linewidth}
        \centering
        \resizebox{1.0\linewidth}{!}{\input{figures/disentanglement/hist_lex_depth_lex_distance_dep_depth_dep_distance_layer_24.pgf}}
    \end{subfigure}
    \caption{Histograms of dimensions selected by dependency and lexical \scalingvec~after joint training \OD{(refer to sect. 4.3?)}. Best in color.}
    \label{fig:dep-lex-separation}
    \vspace{4cm}
\end{figure}

\section{Discussion}

\tomasz{New section}
The introduction of an orthogonal constraint is a core element of our analysis. The constraint assures that no dimension is enhanced or diminished in the transformation and allows interpreting the magnitude of values in the \scalingvec~as the relevance of each dimension for the objectives.

In an \orthoprobe, the sufficient rank of a transformation is learned during the optimization. The rank regularization is a prerequisite to disentangle the information encoded by the probe (\cref{sec:separtation}). 
The natural question is whether such analysis can be performed by reducing the rank of \emph{Structural Probe} with another regularizer and decomposing linear transformation after the optimization. We argue that it is not possible both in joint and separate probing:

\begin{itemize}
    \setlength{\itemsep}{2pt}
    \item In joint probing for multiple tasks: one \scalingvec~is shared for all the tasks. It is not possible to attribute the dimensions to a specific task.
    \item In separate probing for each task: the decomposition leads to different orthogonal matrices. Hence, the dimensions of distinct \scalingvec \emph{s} do not correspond to each other.
\end{itemize}

\subsection{Limitations}


We focus on syntax annotated in Universal Dependencies and lexical hypernymy encoded in WordNet. We do not claim that there is no correlation between syntactic and lexical information in BERT, just that the topologies of those two structures are encoded separately. It is entirely possible that we could find dimensions overlap when probing for syntax and lexicon in differently annotated datasets.

Conversely to \emph{Structural Probes}, our reformulation of the loss 
(in \cref{eqn:orthogonal-depth-probe-loss} and \cref{eqn:orthogonal-distance-probe-loss}) 
is not convex. We thank one of the anonymous ACL reviewers for pointing it out. Nevertheless, we show that despite non-convexity, our \orthoprobe \emph{s}~achieve similar results to \emph{Structural Probes} and are more selective.



\section{Conclusions}
We have expanded structural probing to new types of auxiliary tasks and introduced a new setting, \orthoprobe, in which probes can be optimized jointly. 
We found out that:
\begin{enumerate}
    \item Results of \orthoprobe \emph{s}~are on par with \emph{Standard Structural Probes} on linguistic tasks. \orthoprobe \emph{s}~are less vulnerable to memorization.
    
    \item In addition to syntactic dependencies 
    \orthoprobe \emph{s}~can be efficiently trained to approximate dependency and depth in WordNet hypernymy trees and positional order.
    
    \item \orthoprobe \emph{s}~can be trained jointly for multiple objectives. In most cases, the performance moderately drops, and selectivity increases. The number of parameters decreases in comparison to training many separate probes.
    
    \item Usually, information necessary for each objective is stored in a subspace of relatively low rank (19 - 263). We can further reduce dimensionality by applying sparsity regularization. For a few objectives (e.g., positional depth, dependency depth),
    the information is hugely focal, and the performance can fall markedly when just 25\% randomly selected dimensions are dropped.
    
    \item  We have found that in most of BERT's layers, the subspace encoding linguistic hypernymy is separated from the subspace encoding dependency syntax and subspace encoding word's position.
    
\end{enumerate}

\subsection{Further work}


Our method can be adjusted for multitask and multilingual settings. Following the observation that the orthogonal transformation can map distributions of embeddings in typologically close languages \cite{mikolov-2013-exploiting, vulic-2020-good}. We think that joint training for many languages may be possible by keeping the same \scalingvec~and adding a separate \orthotransf~per language, fulfilling the role of orthogonal mappings. Another leg of research would be analyzing probes for other linguistic structures, for instance, derivation trees.



    


\section*{Acknowledgments}
We thank Ond\v{r}ej Du\v{s}ek, Greg Durrett, and anonymous reviewers of ACL for valuable comments on previous versions of this paper. This work has been supported by grant 18-02196S of the Czech Science Foundation and by grant 338521 of the Charles University Grant Agency.
We have been using language resources and tools developed, stored and distributed by the LINDAT/CLARIAH-CZ project of the Ministry of Education, Youth and Sports of the Czech Republic (project LM2018101).



\bibliographystyle{acl_natbib}
\bibliography{acl2021}

\begin{thebibliography}{35}
\expandafter\ifx\csname natexlab\endcsname\relax\def\natexlab#1{#1}\fi

\bibitem[{Abadi et~al.(2015)Abadi, Agarwal, Barham, Brevdo, Chen, Citro,
  Corrado, Davis, Dean, Devin, Ghemawat, Goodfellow, Harp, Irving, Isard, Jia,
  Jozefowicz, Kaiser, Kudlur, Levenberg, Man\'{e}, Monga, Moore, Murray, Olah,
  Schuster, Shlens, Steiner, Sutskever, Talwar, Tucker, Vanhoucke, Vasudevan,
  Vi\'{e}gas, Vinyals, Warden, Wattenberg, Wicke, Yu, and
  Zheng}]{tensorflow-2015-whitepaper}
Mart\'{\i}n Abadi, Ashish Agarwal, Paul Barham, Eugene Brevdo, Zhifeng Chen,
  Craig Citro, Greg~S. Corrado, Andy Davis, Jeffrey Dean, Matthieu Devin,
  Sanjay Ghemawat, Ian Goodfellow, Andrew Harp, Geoffrey Irving, Michael Isard,
  Yangqing Jia, Rafal Jozefowicz, Lukasz Kaiser, Manjunath Kudlur, Josh
  Levenberg, Dan Man\'{e}, Rajat Monga, Sherry Moore, Derek Murray, Chris Olah,
  Mike Schuster, Jonathon Shlens, Benoit Steiner, Ilya Sutskever, Kunal Talwar,
  Paul Tucker, Vincent Vanhoucke, Vijay Vasudevan, Fernanda Vi\'{e}gas, Oriol
  Vinyals, Pete Warden, Martin Wattenberg, Martin Wicke, Yuan Yu, and Xiaoqiang
  Zheng. 2015.
\newblock \href {http://tensorflow.org/} {{TensorFlow}: Large-scale machine
  learning on heterogeneous systems}.
\newblock Software available from tensorflow.org.

\bibitem[{Arjovsky et~al.(2016)Arjovsky, Shah, and
  Bengio}]{arjovsky-2016-unitary}
Martin Arjovsky, Amar Shah, and Yoshua Bengio. 2016.
\newblock Unitary evolution recurrent neural networks.
\newblock In \emph{International Conference on Machine Learning}, pages
  1120--1128.

\bibitem[{Bansal et~al.(2018)Bansal, Chen, and Wang}]{bansal-2018-can}
Nitin Bansal, Xiaohan Chen, and Zhangyang Wang. 2018.
\newblock \href
  {http://papers.nips.cc/paper/7680-can-we-gain-more-from-orthogonality-regularizations-in-training-deep-networks.pdf}
  {Can {We} {Gain} {More} from {Orthogonality} {Regularizations} in {Training}
  {Deep} {Networks}?}
\newblock In S.~Bengio, H.~Wallach, H.~Larochelle, K.~Grauman, N.~Cesa-Bianchi,
  and R.~Garnett, editors, \emph{Advances in {Neural} {Information}
  {Processing} {Systems} 31}, pages 4261--4271. Curran Associates, Inc.

\bibitem[{Belinkov et~al.(2017)Belinkov, Durrani, Dalvi, Sajjad, and
  Glass}]{belinkov2017neural}
Yonatan Belinkov, Nadir Durrani, Fahim Dalvi, Hassan Sajjad, and James Glass.
  2017.
\newblock \href {https://doi.org/10.18653/v1/P17-1080} {What do neural machine
  translation models learn about morphology?}
\newblock In \emph{Proceedings of the 55th Annual Meeting of the Association
  for Computational Linguistics (Volume 1: Long Papers)}, pages 861--872,
  Vancouver, Canada. Association for Computational Linguistics.

\bibitem[{Belinkov and Glass(2019)}]{belinkov2019analysis}
Yonatan Belinkov and James Glass. 2019.
\newblock \href {https://doi.org/10.1162/tacl\_a\_00254} {Analysis methods in
  neural language processing: A survey}.
\newblock \emph{Transactions of the Association for Computational Linguistics
  (TACL)}, 7:49--72.

\bibitem[{Blevins et~al.(2018)Blevins, Levy, and Zettlemoyer}]{blevins2018deep}
Terra Blevins, Omer Levy, and Luke Zettlemoyer. 2018.
\newblock \href {https://doi.org/10.18653/v1/P18-2003} {Deep {RNN}s encode soft
  hierarchical syntax}.
\newblock In \emph{Proceedings of the 56th Annual Meeting of the Association
  for Computational Linguistics (Volume 2: Short Papers)}, pages 14--19,
  Melbourne, Australia. Association for Computational Linguistics.

\bibitem[{Choromanski et~al.(2020)Choromanski, Likhosherstov, Dohan, Song,
  Gane, Sarlos, Hawkins, Davis, Mohiuddin, Kaiser, Belanger, Colwell, and
  Weller}]{choromanski-2020-rethinking}
Krzysztof Choromanski, Valerii Likhosherstov, David Dohan, Xingyou Song,
  Andreea Gane, Tamas Sarlos, Peter Hawkins, Jared Davis, Afroz Mohiuddin,
  Lukasz Kaiser, David Belanger, Lucy Colwell, and Adrian Weller. 2020.
\newblock \href {http://arxiv.org/abs/2009.14794} {Rethinking attention with
  performers}.

\bibitem[{Dangovski et~al.(2019)Dangovski, Jing, Nakov, Tatalovi\'{c}, and
  Solja\v{c}i\'{c}}]{dangovski-2019-rotational}
Rumen Dangovski, Li~Jing, Preslav Nakov, Mi\'{c}o Tatalovi\'{c}, and Marin
  Solja\v{c}i\'{c}. 2019.
\newblock Rotational unit of memory: a novel representation unit for {RNN}s
  with scalable applications.
\newblock \emph{Transaction of the Association of Computational Linguistics},
  7:121--138.

\bibitem[{Devlin et~al.(2019)Devlin, Chang, Lee, and
  Toutanova}]{devlin2019bert}
Jacob Devlin, Ming-Wei Chang, Kenton Lee, and Kristina Toutanova. 2019.
\newblock Bert: Pre-training of deep bidirectional transformers for language
  understanding.
\newblock In \emph{NAACL-HLT}.

\bibitem[{Hewitt and Liang(2019)}]{hewitt-liang-2019-designing}
John Hewitt and Percy Liang. 2019.
\newblock \href {https://doi.org/10.18653/v1/D19-1275} {Designing and
  interpreting probes with control tasks}.
\newblock In \emph{Proceedings of the 2019 Conference on Empirical Methods in
  Natural Language Processing and the 9th International Joint Conference on
  Natural Language Processing (EMNLP-IJCNLP)}, pages 2733--2743, Hong Kong,
  China. Association for Computational Linguistics.

\bibitem[{Hewitt and Manning(2019)}]{hewitt-manning-2019-structural}
John Hewitt and Christopher~D. Manning. 2019.
\newblock A structural probe for finding syntax in word representations.
\newblock In \emph{NAACL-HLT}.

\bibitem[{Jing et~al.(2017{\natexlab{a}})Jing, Gulcehre, Peurifoy, Shen,
  Tegmark, Soljačić, and Bengio}]{jing-2017-gated}
Li~Jing, Caglar Gulcehre, John Peurifoy, Yichen Shen, Max Tegmark, Marin
  Soljačić, and Y.~Bengio. 2017{\natexlab{a}}.
\newblock \href {https://doi.org/10.1162/neco_a_01174} {Gated orthogonal
  recurrent units: On learning to forget}.
\newblock \emph{Neural Computation}, 31.

\bibitem[{Jing et~al.(2017{\natexlab{b}})Jing, Shen, Dubcek, Peurifoy, Skirlo,
  LeCun, Tegmark, and Solja{\v{c}}i{\'c}}]{jing-2017-eunn}
Li~Jing, Yichen Shen, Tena Dubcek, John Peurifoy, Scott Skirlo, Yann LeCun, Max
  Tegmark, and Marin Solja{\v{c}}i{\'c}. 2017{\natexlab{b}}.
\newblock \href {http://proceedings.mlr.press/v70/jing17a.html} {Tunable
  efficient unitary neural networks ({EUNN}) and their application to {RNN}s}.
\newblock In \emph{Proceedings of the 34th International Conference on Machine
  Learning}, volume~70 of \emph{Proceedings of Machine Learning Research},
  pages 1733--1741, International Convention Centre, Sydney, Australia. PMLR.

\bibitem[{Kingma and Ba(2014)}]{kingma-2014-adam}
Diederik Kingma and Jimmy Ba. 2014.
\newblock Adam: A method for stochastic optimization.
\newblock \emph{International Conference on Learning Representations}.

\bibitem[{Kulmizev et~al.(2020)Kulmizev, Ravishankar, Abdou, and
  Nivre}]{kulmizev-etal-2020-neural}
Artur Kulmizev, Vinit Ravishankar, Mostafa Abdou, and Joakim Nivre. 2020.
\newblock \href {https://doi.org/10.18653/v1/2020.acl-main.375} {Do neural
  language models show preferences for syntactic formalisms?}
\newblock In \emph{Proceedings of the 58th Annual Meeting of the Association
  for Computational Linguistics}, pages 4077--4091, Online. Association for
  Computational Linguistics.

\bibitem[{Linzen et~al.(2016)Linzen, Dupoux, and
  Goldberg}]{linzen2016assessing}
Tal Linzen, Emmanuel Dupoux, and Yoav Goldberg. 2016.
\newblock \href {https://doi.org/10.1162/tacl_a_00115} {Assessing the ability
  of {LSTM}s to learn syntax-sensitive dependencies}.
\newblock \emph{Transactions of the Association for Computational Linguistics},
  4:521--535.

\bibitem[{Liu et~al.(2019)Liu, Gardner, Belinkov, Peters, and
  Smith}]{liu2019linguistic}
Nelson~F. Liu, Matt Gardner, Yonatan Belinkov, Matthew~E. Peters, and Noah~A.
  Smith. 2019.
\newblock Linguistic knowledge and transferability of contextual
  representations.
\newblock In \emph{NAACL-HLT}.

\bibitem[{Mikolov et~al.(2013)Mikolov, Le, and
  Sutskever}]{mikolov-2013-exploiting}
Tomas Mikolov, Quoc~V. Le, and Ilya Sutskever. 2013.
\newblock Exploiting similarities among languages for machine translation.

\bibitem[{Miller(1995)}]{miller-1995-wordnet}
George~A. Miller. 1995.
\newblock \href {https://doi.org/10.1145/219717.219748} {Wordnet: A lexical
  database for english}.
\newblock \emph{Commun. ACM}, 38(11):39–41.

\bibitem[{Nivre et~al.(2020)Nivre, de~Marneffe, Ginter, Haji{\v{c}}, Manning,
  Pyysalo, Schuster, Tyers, and Zeman}]{nivre-etal-2020-ud}
Joakim Nivre, Marie-Catherine de~Marneffe, Filip Ginter, Jan Haji{\v{c}},
  Christopher~D. Manning, Sampo Pyysalo, Sebastian Schuster, Francis Tyers, and
  Daniel Zeman. 2020.
\newblock \href {https://www.aclweb.org/anthology/2020.lrec-1.497} {{U}niversal
  {D}ependencies v2: An evergrowing multilingual treebank collection}.
\newblock In \emph{Proceedings of the 12th Language Resources and Evaluation
  Conference}, pages 4034--4043, Marseille, France. European Language Resources
  Association.

\bibitem[{Peters et~al.(2018)Peters, Neumann, Iyyer, Gardner, Clark, Lee, and
  Zettlemoyer}]{peters2018deep}
Matthew~E. Peters, Mark Neumann, Mohit Iyyer, Matt Gardner, Christopher Clark,
  Kenton Lee, and Luke Zettlemoyer. 2018.
\newblock Deep contextualized word representations.
\newblock In \emph{Proceedings of the 2018 Conference of the North {A}merican
  Chapter of the Association for Computational Linguistics: Human Language
  Technologies, Volume 1 (Long Papers)}, New Orleans, Louisiana. Association
  for Computational Linguistics.

\bibitem[{Pimentel et~al.(2020)Pimentel, Saphra, Williams, and
  Cotterell}]{pimentel-2020-pareto}
Tiago Pimentel, Naomi Saphra, Adina Williams, and Ryan Cotterell. 2020.
\newblock \href {https://doi.org/10.18653/v1/2020.emnlp-main.254} {{P}areto
  probing: {T}rading off accuracy for complexity}.
\newblock In \emph{Proceedings of the 2020 Conference on Empirical Methods in
  Natural Language Processing (EMNLP)}, pages 3138--3153, Online. Association
  for Computational Linguistics.

\bibitem[{Ravichander et~al.(2020)Ravichander, Hovy, Suleman, Trischler, and
  Cheung}]{ravichander-2020-systematicity}
Abhilasha Ravichander, Eduard Hovy, Kaheer Suleman, Adam Trischler, and Jackie
  Chi~Kit Cheung. 2020.
\newblock \href {https://www.aclweb.org/anthology/2020.starsem-1.10} {On the
  systematicity of probing contextualized word representations: The case of
  hypernymy in {BERT}}.
\newblock In \emph{Proceedings of the Ninth Joint Conference on Lexical and
  Computational Semantics}, pages 88--102, Barcelona, Spain (Online).
  Association for Computational Linguistics.

\bibitem[{Rogers et~al.(2020)Rogers, Kovaleva, and
  Rumshisky}]{rogers-2020-primer}
Anna Rogers, Olga Kovaleva, and Anna Rumshisky. 2020.
\newblock \href {https://doi.org/10.1162/tacl_a_00349} {A primer in bertology:
  What we know about how bert works}.
\newblock \emph{Transactions of the Association for Computational Linguistics},
  8:842--866.

\bibitem[{Silveira et~al.(2014)Silveira, Dozat, de~Marneffe, Bowman, Connor,
  Bauer, and Manning}]{silveira-2014-gold}
Natalia Silveira, Timothy Dozat, Marie-Catherine de~Marneffe, Samuel Bowman,
  Miriam Connor, John Bauer, and Christopher~D. Manning. 2014.
\newblock A gold standard dependency corpus for {E}nglish.
\newblock In \emph{Proceedings of the Ninth International Conference on
  Language Resources and Evaluation (LREC-2014)}.

\bibitem[{Student(1908)}]{student-1908-probable}
Student. 1908.
\newblock The probable error of a mean.
\newblock \emph{Biometrika}, pages 1--25.

\bibitem[{Tenney et~al.(2019{\natexlab{a}})Tenney, Das, and
  Pavlick}]{tenney-2019-bert}
Ian Tenney, Dipanjan Das, and Ellie Pavlick. 2019{\natexlab{a}}.
\newblock \href {https://doi.org/10.18653/v1/P19-1452} {{BERT} rediscovers the
  classical {NLP} pipeline}.
\newblock In \emph{Proceedings of the 57th Annual Meeting of the Association
  for Computational Linguistics}, pages 4593--4601, Florence, Italy.
  Association for Computational Linguistics.

\bibitem[{Tenney et~al.(2019{\natexlab{b}})Tenney, Xia, Chen, Wang, Poliak,
  McCoy, Kim, Durme, Bowman, Das, and Pavlick}]{tenney-2018-learn}
Ian Tenney, Patrick Xia, Berlin Chen, Alex Wang, Adam Poliak, R~Thomas McCoy,
  Najoung Kim, Benjamin~Van Durme, Sam Bowman, Dipanjan Das, and Ellie Pavlick.
  2019{\natexlab{b}}.
\newblock \href {https://openreview.net/forum?id=SJzSgnRcKX} {What do you learn
  from context? probing for sentence structure in contextualized word
  representations}.
\newblock In \emph{International Conference on Learning Representations}.

\bibitem[{Torroba~Hennigen et~al.(2020)Torroba~Hennigen, Williams, and
  Cotterell}]{hennigen-2020-intrinsic}
Lucas Torroba~Hennigen, Adina Williams, and Ryan Cotterell. 2020.
\newblock \href {https://doi.org/10.18653/v1/2020.emnlp-main.15} {Intrinsic
  probing through dimension selection}.
\newblock In \emph{Proceedings of the 2020 Conference on Empirical Methods in
  Natural Language Processing (EMNLP)}, pages 197--216, Online. Association for
  Computational Linguistics.

\bibitem[{Vaswani et~al.(2017)Vaswani, Shazeer, Parmar, Uszkoreit, Jones,
  Gomez, Kaiser, and Polosukhin}]{vaswani2017attention}
Ashish Vaswani, Noam Shazeer, Niki Parmar, Jakob Uszkoreit, Llion Jones,
  Aidan~N. Gomez, Lukasz Kaiser, and Illia Polosukhin. 2017.
\newblock Attention is all you need.
\newblock In \emph{Advances in Neural Information Processing Systems 30: Annual
  Conference on Neural Information Processing Systems 2017, 4-9 December 2017,
  Long Beach, CA, {USA}}, pages 5998--6008.

\bibitem[{Vorontsov et~al.(2017)Vorontsov, Trabelsi, Kadoury, and
  Pal}]{vorontsov-2017-orthogonality}
Eugene Vorontsov, Chiheb Trabelsi, Samuel Kadoury, and Chris Pal. 2017.
\newblock \href {http://proceedings.mlr.press/v70/vorontsov17a.html} {On
  orthogonality and learning recurrent networks with long term dependencies}.
\newblock In \emph{Proceedings of the 34th International Conference on Machine
  Learning}, volume~70 of \emph{Proceedings of Machine Learning Research},
  pages 3570--3578, International Convention Centre, Sydney, Australia. PMLR.

\bibitem[{Vuli{\'c} et~al.(2020)Vuli{\'c}, Ruder, and
  S{\o}gaard}]{vulic-2020-good}
Ivan Vuli{\'c}, Sebastian Ruder, and Anders S{\o}gaard. 2020.
\newblock \href {https://doi.org/10.18653/v1/2020.emnlp-main.257} {Are all good
  word vector spaces isomorphic?}
\newblock In \emph{Proceedings of the 2020 Conference on Empirical Methods in
  Natural Language Processing (EMNLP)}, pages 3178--3192, Online. Association
  for Computational Linguistics.

\bibitem[{Wisdom et~al.(2016)Wisdom, Powers, Hershey, Le~Roux, and
  Atlas}]{wisdom-2016-fullcapacity}
Scott Wisdom, Thomas Powers, John Hershey, Jonathan Le~Roux, and Les Atlas.
  2016.
\newblock \href
  {https://proceedings.neurips.cc/paper/2016/file/d9ff90f4000eacd3a6c9cb27f78994cf-Paper.pdf}
  {Full-capacity unitary recurrent neural networks}.
\newblock In \emph{Advances in Neural Information Processing Systems},
  volume~29, pages 4880--4888. Curran Associates, Inc.

\bibitem[{Wolf et~al.(2020)Wolf, Debut, Sanh, Chaumond, Delangue, Moi, Cistac,
  Rault, Louf, Funtowicz, Davison, Shleifer, von Platen, Ma, Jernite, Plu, Xu,
  Scao, Gugger, Drame, Lhoest, and Rush}]{wolf-2020-transformers}
Thomas Wolf, Lysandre Debut, Victor Sanh, Julien Chaumond, Clement Delangue,
  Anthony Moi, Pierric Cistac, Tim Rault, Rémi Louf, Morgan Funtowicz, Joe
  Davison, Sam Shleifer, Patrick von Platen, Clara Ma, Yacine Jernite, Julien
  Plu, Canwen Xu, Teven~Le Scao, Sylvain Gugger, Mariama Drame, Quentin Lhoest,
  and Alexander~M. Rush. 2020.
\newblock \href {https://www.aclweb.org/anthology/2020.emnlp-demos.6}
  {Transformers: State-of-the-art natural language processing}.
\newblock In \emph{Proceedings of the 2020 Conference on Empirical Methods in
  Natural Language Processing: System Demonstrations}, pages 38--45, Online.
  Association for Computational Linguistics.

\bibitem[{Zhang and Bowman(2018)}]{zhang-bowman-2018-language}
Kelly Zhang and Samuel Bowman. 2018.
\newblock \href {https://doi.org/10.18653/v1/W18-5448} {Language modeling
  teaches you more than translation does: Lessons learned through auxiliary
  syntactic task analysis}.
\newblock In \emph{Proceedings of the 2018 {EMNLP} Workshop {B}lackbox{NLP}:
  Analyzing and Interpreting Neural Networks for {NLP}}, pages 359--361,
  Brussels, Belgium. Association for Computational Linguistics.

\end{thebibliography}

\newpage
\appendix

\section{Technical Details}

The \orthoprobe~is trained to minimize L1 loss between predicted and gold distances and depths. The loss is normalized by the number of predictions in a sentence and averaged across a batch of size $12$. Optimization is conducted with Adam \cite{kingma-2014-adam} with initial learning rate $0.02$ and meta parameters: $\beta_1=0.9$, $\beta_2=0.999$, and $\epsilon=10^{-8}$. We use learning rate decay and early-stopping mechanism: if validation loss does not achieve a new minimum after an epoch, learning rate is divided by $10$. After three consecutive learning rate updates not resulting in a new minimum, the training is stopped.

To alleviate sharp jumps in training loss that we observed mainly in training of \emph{Depth Probes}, we clip each gradient's norm at $c=1.5$.

We implemented the network in TensorFlow 2 \cite{tensorflow-2015-whitepaper}. The code is available at GitHub: \url{https://github.com/Tom556/OrthogonalTransformerProbing}.

\subsection{Orthogonal Regularization}

In order to coerce orthogonality of matrix $V$ 
we add DSO to the loss
. \citet{bansal-2018-can} showed that for convolutional neural network applied to image processing, a simpler regularization -- SO is more powerful.

\begin{equation}
    \label{eqn:so}
    \lambda_O SO(V) = \lambda_O||V^TV - \mathbb{I}||^2_F
\end{equation}

\begin{figure}[!h]
    \centering
    \includegraphics[width=0.9\linewidth]{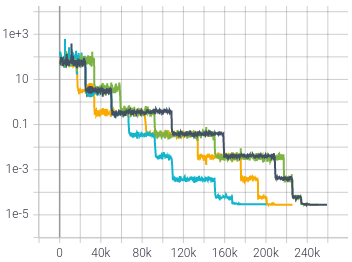}
    \caption{Values of orthogonality penalty during joint training of \orthoprobe~on top of layers: 3 (green), 7 (yellow), 16 (gray), 24 (blue). Optimization steps on the x-axis.}
    \label{fig:orthoreguralization}
\end{figure}

In our experiments, DSO led to faster convergence. \cref{fig:orthoreguralization} shows values of orthogonality penalty during the training. Taking into account the properties of the Frobenius norm, we observe that $V$ matrix is close to orthogonal already after initial epochs.

\subsection{Sparsity Regularization}

\begin{figure}[!h]
    \centering
    \includegraphics[width=0.9\linewidth]{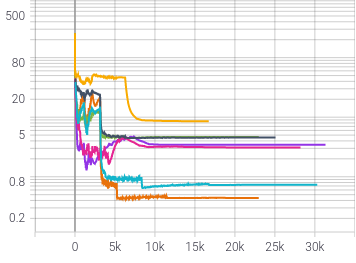}
    \caption{Values of sparsity penalty during separate training of \orthoprobe \emph{s}~with $\lambda=0.05$. Objectives from the highest to the lowest value: lexical distance (yellow), positional distance (green), dependency distance (gray), positional depth (violet), lexical depth (magenta), dependency depth (blue), random depth (orange). Optimization steps on the x-axis.}
    \label{fig:sparsity}
\end{figure}

\cref{fig:sparsity} presents values of sparsity penalty during the training. The regularization is applied only after the orthogonality penalty drops below $1.5$.

\subsection{Number of Parameters}

The number of \orthoprobe \emph{'s} parameters is given by equation:
\begin{equation}
    \label{eqn:DoF}
    NParams_{Ortho} = D_{emb}^2 + D_{emb} \cdot N_{obj},
\end{equation}
where $D_{emb}$ is dimensionality of the embeddings and $N_{obj}$ is a number of jointly probed objectives. Therefore, our biggest probes on top of BERT Large for all eight objectives have $1024^2 + 1024 \cdot 8 = 1,056,768$ parameters. It is more than in \emph{Structural Probes} of \citet{hewitt-manning-2019-structural}. Nevertheless, our probes have less degrees of freedom, because we use \orthotransf~instead of \emph{Linear Transformation}.

\begin{equation}
    DoF_{Ortho} =  \frac{D_{emb} \cdot (D_{emb} -1)}{2} + D_{emb} \cdot N_{obj}
\end{equation}
In the case of joint training for all objectives, the number of degrees of freedom equals to $523,766$.

\subsection{Computation Time}

We have trained \orthoprobe \emph{s}~on GPU a core \textit{GeForce GTX 1080 Ti}. Approximate run times of specific configurations:
\begin{itemize}
    \item separate probing for depth $\sim$ 3 minutes
    \item separate probing for distance $\sim$ 5 minutes
    \item joint probing for distance and depth in the same structure type $\sim$ 7 minutes
    \item joint probing for depths in all structures $\sim$ 13 minutes
    \item joint probing for distance in all structures $\sim$ 18 minutes
    \item probing for all objectives together $\sim$ 35 minutes
\end{itemize}

\section{Derivation of Orthogonal Structural Probe Equation}

\cref{eqn:orthogonal-probe-derivation} with intermediate steps:
\begin{equation}
\begin{split}
    &d_B(h_i,h_j)^2 \\
    &= (U D V^T(h_i - h_j))^T 
    (U D V^T(h_i - h_j)) \\
    &= (h_i - h_j)^T  V  D^T U^T  U  D  V^T  (h_i - h_j) \\
    &= (h_i - h_j)^T  V  D^T D  V^T  (h_i - h_j) \\
    &= (D V^T (h_i - h_j))^T(D V^T (h_i - h_j))
\end{split}
\end{equation}

\section{Dataset Description}

 Universal Dependencies English Web Treebank \citep{silveira-2014-gold} is available at \url{https://github.com/UniversalDependencies/UD_English-EWT}. It consist of: $12,543$ test, $2,002$ dev, and $2,077$ test sentences.

\section{Application in Dependency Parsing}

We have computed the UAS of dependency trees predicted based on dependency probes. We employ the algorithm for extraction of directed dependency trees proposed by \citet{kulmizev-etal-2020-neural}. Our innovation to the method is that we optimize distance and depth probes jointly during one optimization.

\begin{table}[h]
\centering
\begin{tabular}{@{}l|ccc@{}}
\toprule
 Training config. & Layer & UUAS & UAS\\ \midrule
Structural Probe & 15 & 82.29 & -- \\
Orthogonal Probe  & 15 & 82.47 & -- \\ \midrule
\multicolumn{4}{c}{multitask orthogonal probing} \\ \midrule
distance + depth & 16 & 80.86 & 77.51 \\
all distances & 15 & 80.72 & -- \\ 
all tasks & 16 & 79.03 & 75.66 \\\bottomrule
\end{tabular}
\caption{(Undirected) Unlabeled Attachment Score of trees extracted from dependency probes.}
\label{tab:uas}
\end{table}

In line with the previous studies, we show that \orthoprobe \emph{s} can be employed for parsing. \cref{tab:uas} presents Unlabeled Attachment Scores achieved by different multi-task configurations. Joint probing for dependency distance and depth allows us to extract a directed dependency tree in just one optimization. Best to our knowledge, it has not been tried before. Analogically to Spearman's correlation, UAS drops when more objectives are used in optimization. However, even joint probing for all eight objectives is capable of producing trees with $75.66\%$ UAS.
\section{Scaling Vector Properties}

In this appendix, we elaborate on the properties of \scalingvec \emph{s} parameters in the multi-task probing.

\subsection{Parameters Distribution}

\begin{figure}[h]
    \centering
    \resizebox{\linewidth}{!}{\input{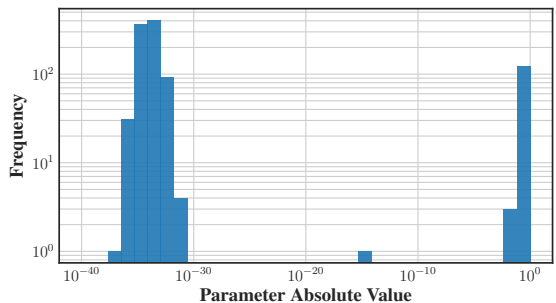}}
    \caption{Logarithmic histogram of \scalingvec~parameters for dependency distance. Joint probing of 16th layer's representations.}
    \label{fig:scalinvec-hist}
\end{figure}

The distribution of values in \scalingvec~(\cref{fig:scalinvec-hist}) shows that the majority of parameters converge to zero. They are within $10^{-40}$ to $10^{-30}$ margin after training. Therefore, the significant dimensions are clearly identifiable.

\subsection{Separation of Information (Continued)}
\label{sec:dim-selection-ctnd}

On the following pages, we present dimension overlap histograms and tables, as in \cref{sec:separtation}, for the remaining pairs of objectives.

\begin{figure}[!ht]
    \centering
    \begin{subfigure}{0.9\linewidth}
        \centering
        \resizebox{1.0\linewidth}{!}{\input{figures/disentanglement/hist_dep_depth_dep_distance_pos_depth_pos_distance_layer_1.pgf}}
    \end{subfigure}
    \begin{subfigure}{0.9\linewidth}
        \centering
        \resizebox{1.0\linewidth}{!}{\input{figures/disentanglement/hist_dep_depth_dep_distance_pos_depth_pos_distance_layer_6.pgf}}
    \end{subfigure}
        \begin{subfigure}{0.9\linewidth}
        \centering
        \resizebox{1.0\linewidth}{!}{\input{figures/disentanglement/hist_dep_depth_dep_distance_pos_depth_pos_distance_layer_16.pgf}}
    \end{subfigure}
        \begin{subfigure}{0.9\linewidth}
        \centering
        \resizebox{1.0\linewidth}{!}{\input{figures/disentanglement/hist_dep_depth_dep_distance_pos_depth_pos_distance_layer_24.pgf}}
    \end{subfigure}
\end{figure}

\begin{table}[!h]
\centering
\tiny
    \begin{tabular}{ll|cc|cc|cc|cc}
    & & \multicolumn{2}{c}{DEP} & \multicolumn{2}{|c}{LEX} & \multicolumn{2}{|c}{POS} & \multicolumn{2}{|c}{RAND} \\
    & & \rotatebox{90}{Depth} & \rotatebox{90}{Dist.} & \rotatebox{90}{Depth} & \rotatebox{90}{Dist.} & \rotatebox{90}{Depth} & \rotatebox{90}{Dist.} & \rotatebox{90}{Depth} & \rotatebox{90}{Dist.} \\\hline
    \multirow{5}{*}{\rotatebox{90}{DEP}} &  &   &   &  &  &  &  & \\
     & Depth & 65 & 54 & 1 &  18 & 11 & 24 & 48 & 44 \\
    &  &   &   &  &  &  &  & \\
     & Dist. &  & 109 &6 &  43 & 11 & 39 & 45 & 64  \\ 
    &  &   &   &  &  &  &  & \\ \hline
    \multirow{5}{*}{\rotatebox{90}{LEX}} &  &   &   &  &  &  &  & \\
     & Depth &  &  & 46  & 45 & 2 &  13 & 7 &  8  \\
    &  &   &   &  &  &  &  & \\
     & Dist. &  &  &  & 551 &2 &  42 & 46 & 103  \\ 
    &  &   &   &  &  &  &  & \\ \hline
    \multirow{5}{*}{\rotatebox{90}{POS}} &  &   &   &  &  &  &  & \\
     & Depth &  &  &  &   &  20 & 11 & 20 & 14 \\
    &  &   &   &  &  &  &  & \\
     & Dist. &  &  &  &  &  & 111 & 47 & 71  \\
    &  &   &   &  &  &  &  & \\ \hline
    \multirow{5}{*}{\rotatebox{90}{RAND}} &  &   &   &  &  &  &  & \\
     & Depth &  &  &  &  &  &  & 152 & 112  \\
    &  &   &   &  &  &  &  & \\
     & Dist. &  &  &  &  &  &  & & 265 \\
    &  &   &   &  &  &  &  & \\ \hline
    \end{tabular}
    \caption{Number of shared dimensions selected by \scalingvec~after the joint training of probe on top of the 1st layer.}
    \label{tab:l-1-separation}
\end{table}

\begin{figure}[!ht]
    \centering
    \begin{subfigure}{0.9\linewidth}
        \centering
        \resizebox{1.0\linewidth}{!}{\input{figures/disentanglement/hist_lex_depth_lex_distance_pos_depth_pos_distance_layer_1.pgf}}
    \end{subfigure}
    \begin{subfigure}{0.9\linewidth}
        \centering
        \resizebox{1.0\linewidth}{!}{\input{figures/disentanglement/hist_lex_depth_lex_distance_pos_depth_pos_distance_layer_6.pgf}}
    \end{subfigure}
        \begin{subfigure}{0.9\linewidth}
        \centering
        \resizebox{1.0\linewidth}{!}{\input{figures/disentanglement/hist_lex_depth_lex_distance_pos_depth_pos_distance_layer_16.pgf}}
    \end{subfigure}
        \begin{subfigure}{0.9\linewidth}
        \centering
        \resizebox{1.0\linewidth}{!}{\input{figures/disentanglement/hist_lex_depth_lex_distance_pos_depth_pos_distance_layer_24.pgf}}
    \end{subfigure}
\end{figure}

\begin{table}[!h]
\centering
\tiny
    \begin{tabular}{ll|cc|cc|cc|cc}
    & & \multicolumn{2}{c}{DEP} & \multicolumn{2}{|c}{LEX} & \multicolumn{2}{|c}{POS} & \multicolumn{2}{|c}{RAND} \\
    & & \rotatebox{90}{Depth} & \rotatebox{90}{Dist.} & \rotatebox{90}{Depth} & \rotatebox{90}{Dist.} & \rotatebox{90}{Depth} & \rotatebox{90}{Dist.} & \rotatebox{90}{Depth} & \rotatebox{90}{Dist.} \\\hline
    \multirow{5}{*}{\rotatebox{90}{DEP}} &  &   &   &  &  &  &  & \\
     & Depth & 50 & 43 & 0 &  1 &  11 & 26 & 30 & 26 \\
    &  &   &   &  &  &  &  & \\
     & Dist. &  & 81 & 1 &  2 &  11 & 38 & 35 & 39  \\ 
    &  &   &   &  &  &  &  & \\ \hline
    \multirow{5}{*}{\rotatebox{90}{LEX}} &  &   &   &  &  &  &  & \\
     & Depth & &  &  30 & 28 & 0 &  4 &  1 &  6  \\
    &  &   &   &  &  &  &  & \\
     & Dist. &  &  &   & 346 &0 &  19 & 14 & 45  \\ 
    &  &   &   &  &  &  &  & \\ \hline
    \multirow{5}{*}{\rotatebox{90}{POS}} &  &   &   &  &  &  &  & \\
     & Depth & &  & &  &  14 & 11 & 13 & 11 \\
    &  &   &   &  &  &  &  & \\
     & Dist. &  &  & &   & & 99 & 41 & 71  \\
    &  &   &   &  &  &  &  & \\ \hline
    \multirow{5}{*}{\rotatebox{90}{RAND}} &  &   &   &  &  &  &  & \\
     & Depth &  &  &  &   &  &  & 113 & 70  \\
    &  &   &   &  &  &  &  & \\
     & Dist. &  & &  &   &  &  &  & 267 \\
    &  &   &   &  &  &  &  & \\ \hline
    \end{tabular}
    \caption{Number of shared dimensions selected by \scalingvec~after the joint training of probe on top of the 6th layer.}
    \label{tab:l-6-separation}
\end{table}

\begin{figure}[!ht]
    \centering
    \begin{subfigure}{0.9\linewidth}
        \centering
        \resizebox{1.0\linewidth}{!}{\input{figures/disentanglement/hist_dep_depth_dep_distance_rnd_depth_rnd_distance_layer_1.pgf}}
    \end{subfigure}
    \begin{subfigure}{0.9\linewidth}
        \centering
        \resizebox{1.0\linewidth}{!}{\input{figures/disentanglement/hist_dep_depth_dep_distance_rnd_depth_rnd_distance_layer_6.pgf}}
    \end{subfigure}
        \begin{subfigure}{0.9\linewidth}
        \centering
        \resizebox{1.0\linewidth}{!}{\input{figures/disentanglement/hist_dep_depth_dep_distance_rnd_depth_rnd_distance_layer_16.pgf}}
    \end{subfigure}
        \begin{subfigure}{0.9\linewidth}
        \centering
        \resizebox{1.0\linewidth}{!}{\input{figures/disentanglement/hist_dep_depth_dep_distance_rnd_depth_rnd_distance_layer_24.pgf}}
    \end{subfigure}
\end{figure}

\begin{table}[!h]
\centering
\tiny
    \begin{tabular}{ll|cc|cc|cc|cc}
    & & \multicolumn{2}{c}{DEP} & \multicolumn{2}{|c}{LEX} & \multicolumn{2}{|c}{POS} & \multicolumn{2}{|c}{RAND} \\
    & & \rotatebox{90}{Depth} & \rotatebox{90}{Dist.} & \rotatebox{90}{Depth} & \rotatebox{90}{Dist.} & \rotatebox{90}{Depth} & \rotatebox{90}{Dist.} & \rotatebox{90}{Depth} & \rotatebox{90}{Dist.} \\\hline
    \multirow{5}{*}{\rotatebox{90}{DEP}} &  &   &   &  &  &  &  & \\
     & Depth & 189 & 144 & 17 & 39 & 70 & 66 & 146 & 123 \\
    &  &   &   &  &  &  &  & \\
     & Dist. &  & 463 & 16 & 82 & 81 & 141 & 186 & 275  \\ 
    &  &   &   &  &  &  &  & \\ \hline
    \multirow{5}{*}{\rotatebox{90}{LEX}} &  &   &   &  &  &  &  & \\
     & Depth &  &  & 33 & 22 & 9 & 10 & 18 & 16  \\
    &  &   &   &  &  &  &  & \\
     & Dist. &  & & & 173 & 30 & 48 & 64 & 98  \\ 
    &  &   &   &  &  &  &  & \\ \hline
    \multirow{5}{*}{\rotatebox{90}{POS}} &  &   &   &  &  &  &  & \\
     & Depth &  &  &   &  & 124 & 70 & 107 & 97 \\
    &  &   &   &  &  &  &  & \\
     & Dist. &  &  & &  &  & 190 & 136 & 177  \\
    &  &   &   &  &  &  &  & \\ \hline
    \multirow{5}{*}{\rotatebox{90}{RAND}} &  &   &   &  &  &  &  & \\
     & Depth &  & &  & &  &  & 287 & 198  \\
    &  &   &   &  &  &  &  & \\
     & Dist. &  &  &  &  &  &  &  & 410 \\
    &  &   &   &  &  &  &  & \\ \hline
    \end{tabular}
    \caption{Number of shared dimensions selected by \scalingvec~after the joint training of probe on top of the 24th layer.}
    \label{tab:l-24-separation}
\end{table}

\begin{figure}[t]
    \centering
    \begin{subfigure}{0.9\linewidth}
        \centering
        \resizebox{1.0\linewidth}{!}{\input{figures/disentanglement/hist_lex_depth_lex_distance_rnd_depth_rnd_distance_layer_1.pgf}}
    \end{subfigure}
    \begin{subfigure}{0.9\linewidth}
        \centering
        \resizebox{1.0\linewidth}{!}{\input{figures/disentanglement/hist_lex_depth_lex_distance_rnd_depth_rnd_distance_layer_6.pgf}}
    \end{subfigure}
        \begin{subfigure}{0.9\linewidth}
        \centering
        \resizebox{1.0\linewidth}{!}{\input{figures/disentanglement/hist_lex_depth_lex_distance_rnd_depth_rnd_distance_layer_16.pgf}}
    \end{subfigure}
        \begin{subfigure}{0.9\linewidth}
        \centering
        \resizebox{1.0\linewidth}{!}{\input{figures/disentanglement/hist_lex_depth_lex_distance_rnd_depth_rnd_distance_layer_24.pgf}}
    \end{subfigure}
\end{figure}

\end{document}